\documentclass[conference]{IEEEtran}
\IEEEoverridecommandlockouts
\usepackage{cite}
\usepackage{amsmath,amssymb,amsfonts}
\usepackage{algorithmic}
\usepackage{graphicx}
\usepackage{textcomp}
\usepackage{xcolor}
\usepackage{url}
\usepackage{balance}
\usepackage{multirow}
\usepackage{multicol}
\usepackage{hhline}

\def\BibTeX{{\rm B\kern-.05em{\sc i\kern-.025em b}\kern-.08em
    T\kern-.1667em\lower.7ex\hbox{E}\kern-.125emX}}
\begin{document}

\title{Detection of Fake Generated Scientific Abstracts\\

}
\author{
\IEEEauthorblockN{Panagiotis C. Theocharopoulos\IEEEauthorrefmark{1}, 
Panagiotis Anagnostou\IEEEauthorrefmark{1},
Anastasia Tsoukala\IEEEauthorrefmark{1},\\
Spiros V. Georgakopoulos\IEEEauthorrefmark{2},
Sotiris K. Tasoulis\IEEEauthorrefmark{1} and
Vassilis P. Plagianakos\IEEEauthorrefmark{1}}
\IEEEauthorblockA{
\IEEEauthorrefmark{1}Department of Computer Science and Biomedical Informatics\\
University of Thessaly, Greece\\
Email: \{ptheochar, panagno, antsoukala, stasoulis, vpp\}@uth.gr}
\IEEEauthorblockA{\IEEEauthorrefmark{2}Department of Mathematics\\
University of Thessaly, Greece\\
Email: spirosgeorg@uth.gr}
}

\maketitle

\begin{abstract}
The widespread adoption of Large Language Models and publicly available ChatGPT has marked a significant turning point in the integration of Artificial Intelligence into people's everyday lives. The academic community has taken notice of these technological advancements and has expressed concerns regarding the difficulty of discriminating between what is real and what is artificially generated. Thus, researchers have been working on developing effective systems to identify machine-generated text. In this study, we utilize the GPT-3 model to generate scientific paper abstracts through Artificial Intelligence and explore various text representation methods when combined with Machine Learning models with the aim of identifying machine-written text. We analyze the models' performance and address several research questions that rise during the analysis of the results. By conducting this research, we shed light on the capabilities and limitations of Artificial Intelligence generated text.  
\end{abstract}


\section{Introduction}
The development of the transformer architecture along with the attention mechanism took the lead on Natural Language Processing (NLP) tasks, due to its ability to process an entire input of sequence at once, unlike the Recurrent Neural Networks (RNNs). The wider use of the transformers architecture leads to Large Language models (LLMs). The LLMs are language models trained on a tremendous amount of data with multiple parameters. Furthermore, the development of pre-trained models allowed language models to learn generic language patterns and structures from large and diverse data before being fine-tuned for a specific task \cite{KASNECI2023102274}. 

After the conversational Artificial Intelligence (AI) tool ChatGPT became publicly available in December 2022, the discipline of computer science, among other fields, was shaken to its core. Developed by OpenAI, ChatGPT is a language model representing a refined and evolved version of Generative Pre-trained Transformer (GPT) models. The tool's remarkable potential for executing a vast array of tasks with high precision has attracted widespread attention.

The introduction of ChatGPT, and similar language models, has been met with excitement and concern from scientists and researchers. Despite the impressive development of language models in AI and NLP, there are valid concerns regarding their misuse~\cite{jawahar2020automatic}. These models may be used for more sophisticated phishing or social engineering schemes and for impersonating individuals or organizations. It is crucial to cautiously protect potentially affected areas by malicious intent and consider the ethical implications with further research~\cite{ma2023abstract}. Journalism, cybersecurity, intellectual property, customer service, and education are most prominently affected by the proliferation of generated text applications. The area of concern of our study is education, as the ability of language models to generate academic writing could lead to a rise in plagiarism and academic dishonesty. Recently, some examples of generated scientific and academic texts have already ended up as conference papers~\cite{crothers2022machine}. The first scientific paper generator was the SCIgen program, created in 2005 by three MIT graduate students.
SCIgen is a computer program, created as a prank, which generates random computer science research papers~\cite{stribling2005scigen}. Detecting synthetic text is getting extremely challenging because sophisticated text generation has been pursued for many years, culminating in the impressive capabilities of GPT-3, which can even deceive human readers~\cite{ippolito2019automatic}. Many text generation models followed either for severe or humorous causes. However, the necessity for a distinction between generated and actual text became even more substantial~\cite{ippolito2019automatic,crothers2022machine}. Currently, GPT-3 has approximately 570 GB of text data as a training dataset, combining the 175 billion parameters~\cite{brown2020language}. Although ChatGPT and GPT-3 differ in terms of training methodology and fine-tuning, the latter is provided via its platform by OpenAI primarily for research purposes~\cite{openai_ind,brown2020language}.

The scope of this study is to investigate a way that distinguishes text generated by LLMs, in the particular scientific text produced by GPT-3. Additionally, the study tries to understand the results by answering logical questions, based on the false outcomes of the best-performed model. The rest of this paper is structured into different sections. Section~\ref{rel_work} provides the related literature about the topic. Section~\ref{method} describes the methodology applied in the study. Section~\ref{discussion}, discusses our thoughts on the findings. We conclude the paper in Section~\ref{conclusion} and suggest ideas for future research directions.

\section{Related Work}\label{rel_work}

Although the concept of AI text recognition based on Large Language Models (LLMs) is relatively new, there is already related research in the field. A tool called Giant Language Model Test Room (GLTR)~\cite{gehrmann2019gltr} is implemented to detect whether a text has been generated by a machine or written by a human. The work is based on several statistical methods to detect AI-generated text from GPT-2 and BERT, including computing the model density of generated output and comparing it to human-generated text, as well as using the probability of a word and the distribution over four buckets of absolute ranks of predictions. On the same page, in order to discriminate machine-generated text, the study \cite{zhong2020neural} proposes the FActual Structure of Text (FAST) method, a graph-based model that utilizes the factual structure of a document for deep fake detection of text. The graph nodes are extracted by Named Entity Recognition (NER). Furthermore, the sentence representations were constructed via document-level aggregation for the final prediction, where the consistency and coherence of continuous sentences are sequentially modeled. For the evaluation of FAST, the authors used AI-generated text data generated from GROVER and GPT-2. The approach outperformed transformer-based models~\cite{zhong2020neural}. Based on the text generated by GPT-2 model, OpenAI created a detector of whether a text is artificially generated. The detector has been based on a fine-tuning version of the RoBERTa base. The model has been trained using $510$ tokens and using $5,000$ text samples from the WebText dataset $5,000$ texts generated by a GPT-2 model. The model achieved $95\%$ of accuracy~\cite{solaiman2019release}. Furthermore, in~\cite{liu2022coco} is presented a coherence-based contrastive learning model (CoCo) for the detection of text generated from machines. To achieve that, the authors modeled the text coherence with entity consistency and sentence interaction. The CoCo model outperformed in terms of accuracy and F1 score similar models such as GPT-2, RoBERTa, and XLNet but also the GROVER and FAST models. Additionally, the authors of \cite{mitchell2023detectgpt} proposed a tool, called DetectGPT, that detects machine-generated text from LLMs. DetectGPT is a more effective approach for detecting machine-generated text than existing zero-shot methods. The authors found that DetectGPT significantly improved the detection of fake news articles generated by $20$B parameter GPT-NeoX, achieving an AUROC of $0.95$ compared to the strongest zero-shot baseline's AUROC of $0.81$. Additionally, the authors found that DetectGPT does not require training a separate classifier or collecting a dataset of a real or generated corpus, making it a more efficient and practical approach for detecting machine-generated text~\cite{mitchell2023detectgpt}.

The \cite{ma2023ai} has focused on the involvement of AI writing in scientific writing. The authors analyze the similarities and differences between the two types of content, including writing style, consistency, coherence, language redundancy, and factual errors. The results suggest that while AI has the potential to generate accurate scientific content, there is still a gap in terms of depth and overall quality. The study utilizes the logistic regression model as a detector along with OpenAI's RoBERTa detector. The results suggest that there are significant differences in the distribution of text between human-written and AI-generated scientific work and that AI-generated abstracts tend to be more generic than human-written ones. Additionally, the logistic regression model achieved a higher F1-score for detecting AI-generated text~\cite{ma2023ai}. 

Finally, as the extensive use of GPT-3 and ChatGPT continues to grow, OpenAI published on January 31, 2023, an update of the aforementioned model. The revised version of the AI text classifier fine-tuned pre-trained models in order to distinguish the AI-generated text. The model used data from three origins of the human-written text, a dataset from Wikipedia, the 2019 WebText dataset, and a group of human examples gathered during the instruction of InstructGPT. Although, the model needs $1,000$ tokens and more in order to produce reliable results. The revised edition of the model has shown better results than the previous version, which was based on GPT-2. In the validation set, the AUC score increased from $0.95$ to $0.97$, and in the challenge set, it increased from $0.43$ to $0.66$. Furthermore, the classifier accurately identifies $26\%$ of text generated by AI as possibly AI-written (true positives) but mistakenly identifies $9\%$ of human-written text as AI-generated (false positives)~\cite{openai_ind}.

\section{Research Methodology}\label{method}
\begin{figure*}[!t]
    \centering
    \includegraphics[width=\linewidth]{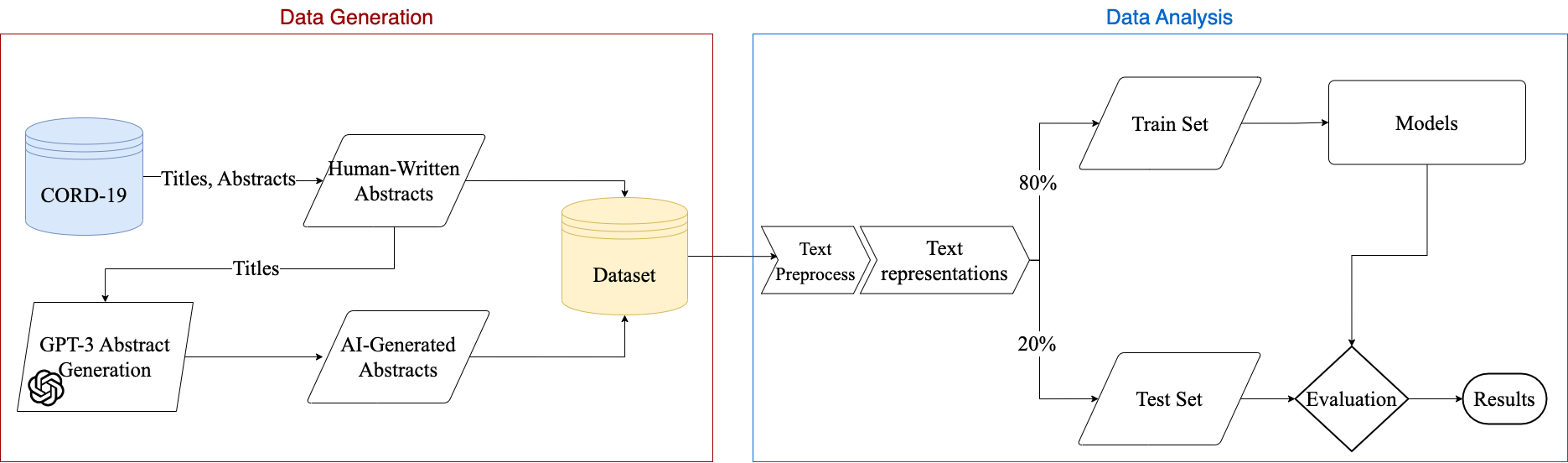}
    \caption{Schematic overview of the study. Dataset Generation: From the CORD-19 data has been collected the titles and the abstracts of the academic literature. The titles of the selected work have been prompted to GPT-3 model, via its API for the AI-generated abstract based on its title (Left). Data Analysis: The study involved text cleaning and data representations using various methods, as well as the models' results evaluation (Right).}
    \label{fig:overview}
\end{figure*}

The purpose of this study is to suggest a strategy for distinguishing between scientific literature produced by humans and AI-generated. We examine various techniques for text representation and Machine Learning models which are applied to both human-written and AI-generated scientific abstracts. We utilized the GPT-3 model to create a database of AI-generated abstracts, which is publicly available\footnote{https://github.com/panagiotisanagnostou/AI-GA}. We examine well-established NLP methods for text representation as well as embedded representation methods in order to train classic ML and Deep Learning models examining their performances and trying to answer three research questions:
\begin{enumerate}
    \item ``\textit{What words dominate the misclassified AI generated texts compared to the correctly classified ones?}''
    \item ``\textit{What is the reason behind the misclassified texts?}''
    \item ``\textit{Does the size of the title we provided the GPT-3 model affect the classification error?}''
\end{enumerate}
The strategy that produces the most accurate outcomes will be recommended for addressing this particular issue. A workflow diagram of the adopted strategy for creating the AI-generated dataset is presented in \figurename~\ref{fig:overview}.

\subsection{AI-Generated Dataset}
For this study, we used the publicly available COVID-19 Open Research Dataset (CORD-19)~\cite{wang2020cord}. The CORD-19 is an aggregation of published papers and pre-prints from multiple sources, which has been collected in order to  promote research on COVID-19 and related coronaviruses. The dataset contains tens of thousands of scholarly articles, some of them with full text. The primary objective of CORD-19 is to provide the global research community with a wealth of information, to support the fight against COVID-19. This has been achieved by creating a freely available dataset, which can be used in conjunction with advanced Natural Language Processing (NLP) and other Artificial Intelligence (AI) techniques to extract new insights. As there is a large number of academic papers regarding coronaviruses, there is a pressing need for text mining and information retrieval systems that can help medical researchers keep up. We randomly selected a subset of academic works from the CORD-19 dataset, consisting of $14,331$ English-language papers with both titles and abstracts.

In order to produce the AI-generated abstracts, we used one of the most advanced LLMs available, the GPT-3. Specifically, we used the GPT-3 model also known as the Davinci model, which is one of the most capable GPT-3 models and produces higher quality results~\cite{openai_API}. Although GPT-3.5 models would have been preferable for this task, they were not publicly available via an API at the time of this experiment. To create the AI abstracts, we developed an appropriate prompt with the following form: \textit{``Create an abstract for a scientific journal with a formal tone, academic language, and a background story of the topic in a unique paragraph with the title: $\mathbf{\hat{t}}$''}, where $\mathbf{\hat{t}}$ is the human-written original title of the academic paper. In order to make the generated text more creative and novel, several adjustments were made to the requested parameters. These adjustments included increasing the freedom to create novelty by selecting the most likely tokens through changes to the randomness of the generated text and penalizing the frequency of repeated words or phrases to generate more diverse responses. Additionally, the model was set to avoid using words or phrases already present in the prompt or previously generated text in order to generate more unique responses. The titles provided for the prompt were the same as the titles selected from the CORD-19 dataset. The final dataset contains $28,662$ entries, involving $14,331$ human-written abstracts and $14,331$ AI-generated abstracts, corresponding to the same titles. In order to ensure that the analysis is based on high-quality data, we had to perform a text-cleaning procedure. The cleaning procedure involved:
\begin{itemize}
    \item Removing special characters such as HTML tags or special characters e.g. next line symbol,
    \item Removing whitespace or line breaks in the text,
    \item Removing stop words, such as ``the'', ``and'', ``of'', etc.\ because they do not convey meaningful information,
    \item Removing any non-alphabetic characters or numbers that may be present in the text,
    \item Normalizing text involves converting all text to a consistent format, such as lowercase, to make it easier to process and analyze,
    \item Removing the two most frequent words, namely ``paper'' and ``study'' from both original and generated texts due to their high frequency on the generated text.
\end{itemize}

\subsection{Text representation}\label{text_representation}

Text representation is an essential step in NLP. The main purpose is to transform unstructured text data into a structured and machine-readable format. The purpose of text representation is to enable text algorithms to perform various NLP tasks such as text classification, sentiment analysis, and machine translation. In the following section, we describe some of the methods used as we carried out our tests. 

\subsubsection{Term Frequency-Inverse Document Frequency (TF-IDF)}
TF-IDF is a method employed in NLP and information retrieval to evaluate the significance of a term in a document or corpus. It is created by combining two distinct components: Term Frequency (TF) and Inverse Document Frequency (IDF). The TF component quantifies the number of occurrences of a term in a document, while the IDF component assigns less weight to commonly used words and more weight to less frequent words. When a word appears more frequently in documents, it will have a higher term frequency, while less frequent occurrences of a word will result in greater importance (IDF) for that word when it is searched in a specific document. TF-IDF is the product of multiplying TF and IDF components~\cite{qaiser2018text}. During our study, we removed words that exhibited less than 1\% or more than 99\% of their values to filter out unwanted noise in the data from sparse and dense variables.

\subsubsection{Named Entity Recognition}
Named Entity Recognition (NER) is a fundamental task in the field of NLP, aimed at identifying and classifying named entities from a given text. The task of NER involves analyzing text data and identifying named entities based on a range of linguistic features, including part-of-speech tags, syntactic structures, and context. NER algorithms employ various machine learning techniques such as statistical models, neural networks, and rule-based methods to automatically detect and extract named entities from unstructured text data. The output of a NER system is typically a structured representation of the identified entities, which may include labels indicating their types~\cite{goyal2018recent, Neumann_2019}. As we conducted our experiments, we utilized a bioinformatics-focused approach due to the main topic of the abstracts being about COVID-19, employing NER via the SciSpacy python package's {``en\_core\_sci\_lg''} model to identify relevant bioinformatics terminology~\cite{Neumann_2019}. The extracted n-grams, up to a size of 9, were then used as input for our classification models. 

\subsubsection{Word2Vec}
Word2Vec is a popular method for generating word embeddings, which are vector representations of words that capture their semantic and syntactic meaning~\cite{mikolov2013efficient}.
Word2Vec is a widely used natural language processing technique that is predicated on the premise that words that co-occur in similar linguistic contexts tend to have similar semantic meanings. It comprises two neural network-based models, namely the Continuous Bag of Words (CBOW) and Skip-gram models, which generate word embeddings from large text corpora. These embeddings encode the semantic and syntactic relationships between words, thereby facilitating a range of natural language processing applications\cite{mikolov2013efficient, goldberg2014word2vec}. In this experiment, we used the pre-trained model google-news-300, an unsupervised natural language processing model developed by Google. It is based on the Word2Vec technique and is designed to generate high-quality word embeddings from large corpora of news articles. The model consists of 300-dimensional vectors, each of which represents a unique word in the vocabulary.

\subsubsection{Contextualized Representations}
Contextualized representations refer to the capability of language models to produce unique and dynamic representations of words and sentences based on the surrounding context. These representations not only capture the inherent meaning of individual words but also the subtleties and nuances of their usage within a specific context. The use of contextualized representations enhances the accuracy and naturalness of language output and is increasingly vital in NLP tasks such as sentiment analysis, question answering, and machine translation. Contextualized representations are achieved through deep neural network architectures that are trained on extensive text data. These models usually rely on recurrent or transformer-based architectures that can capture long-range dependencies between words and phrases in a text~\cite{ethayarajh2019contextual}.

An example of such a model is the Bidirectional Encoder Representations from Transformers (BERT), which has achieved state-of-the-art results on a variety of natural language processing tasks~\cite{devlin2018bert}. BERT is a type of machine learning model that uses bidirectional representation, which allows it to comprehend the entire sequence of words in the context of the sentence. The model is trained using a Masked Language Model (MLM) objective, which involves masking out $15\%$ of the input words during training. These masked tokens are replaced with either a ``MASK'' token, a random word, or the same word, with frequencies of $80\%$, $10\%$, and $10\%$, respectively~\cite{chintalapudi2021sentimental}. BERT comes in two models: the base model with 12 encoders and the large model with 64 encoders.
Additionally, the BERT model has the ability to perform Next Sentence Prediction, where it is pre-trained on pairs of text to learn the relationships between sentences and determine if a given sentence follows the previous sentence or not. BERT's input includes Token Embeddings, Segment Embeddings, and Positional Embeddings. The word tokenization process is performed by the BERT tokenizer, which uses the concept of word-piece tokenizer to break down certain words into sub-words or word pieces if the word can be represented by multiple tokens~\cite{devlin2018bert,theocharopoulos2022text,chintalapudi2021sentimental}.
During our research experiments, we used the BERT base pre-trained model. On top of BERT’s architecture, we added a fully connected layer, where each neuron is connected to every neuron in the previous layer, and each connection has an associated weight parameter that is learned during training. In addition, a regularization technique randomly drops out the $50\%$ of the neurons in the layer during each training iteration, to prevent the model from overfitting over the training data.

In our experiments, we adopted an alternative methodology that involved the use of a pre-trained NLP model designed for general usage. The objective behind this approach was to map our data onto the embedding space established by the pre-existing model. Specifically, we employed the ``English Wikipedia Dump of November 2021'' model to achieve this goal~\cite{kutuzov2017word}. The Wikipedia dataset is comprised of filtered articles from all languages. The construction of these datasets entails the extraction of data from the Wikipedia dump, with each language having a distinct file. Each record within the dataset corresponds to a complete Wikipedia article, which has been processed to exclude superfluous sections, such as markdown, references, and other unwanted content~\cite{wikidump}. By adopting this technique, we succeeded in creating a data matrix of $300$ variables.

\section{Experimental Results}\label{results}
This section presents the results of how each method of word representation technique, combined with the machine learning methods, performed and distinguish the original abstracts from the AI-generated ones. To evaluate the effectiveness of the methods, we conducted a comparison using a range of text representation techniques as mentioned in Section~\ref{text_representation} with various Machine Learning (ML) methods such as Logistic Regression (LR), Multinomial Naive Bayes (MNB), Support Vector Machine (SVM), Long Short-Term Memory networks (LSTM), and BERT.

Our analysis involved training and testing these algorithms on a large corpus of original and AI-generated abstracts. The training of the methods has involved the $80\%$ of the corpus, containing $22,930$ entries, and the evaluation of the remaining $20\%$ of $5,732$ entries. Furthermore, the training and the evaluation entries contain both the real and the generated abstract from each of the paper titles we used in the abstract generation process. For the evaluation of the performance of each algorithm, we used the metrics such as accuracy, precision, recall, and F1 score.

The combination of text representations with the ML models can be summarized as follows: 
\begin{itemize}
    \item LR with TF-IDF
    \item MNB with TF-IDF 
    \item SVM with TF-IDF
    \item LR with NER 
    \item MNB with NER 
    \item SVM with NER
    \item LR with EWD 
    \item SVM with EWD 
    \item BERT with keeping the initial weights fixed
    \item BERT with fine-tuning the model by updating the initial weights (BERT Fine-Tuning)
    \item LSTM with BERT
    \item LSTM with Word2Vec
\end{itemize}

Table~\ref{tab:results} presents the average performance of each method over 100 independent iterations. To mitigate the risk of overfitting, we employed an early stopping strategy by continuously monitoring the train and validation loss, particularly for the LSTM and BERT models. Specifically, once the validation loss reached its minimum value and began to increase, we terminated the training process and saved the current model. Our analysis reveals that the LSTM model combined with Word2Vec representation achieved the highest accuracy rate of $98.7\%$. Although the BERT model performed equally well in terms of accuracy and better in F1 score, a composite measure of precision and recall, it exhibited lower precision and recall scores. The LSTM model incorporating Word2Vec achieved an AUC score of 0.987, which is higher than the AUC scores of the original and updated OpenAI models. The OpenAI models were developed to distinguish AI-generated text from human-written text, and their AUC scores were 0.95 and 0.97, respectively. 

\begin{table}[!t]
\caption{Model results}
\label{tab:results}
\begin{center}
\begin{tabular}{|l|llll|}
\hline
\multirow{2}{*}{Models} & \multicolumn{4}{c|}{Metrics} \\ \cline{2-5} 
                        & Accuracy & F1 & Precision & Recall \\ \hline
LSTM + BERT             & 0.697    & 0.700    & 0.692      & 0.708 \\ \hline
MNB + NER               & 0.797    & 0.800    & 0.811      & 0.789 \\ \hline
BERT                    & 0.834    & 0.830    & 0.830      & 0.830 \\ \hline
SVM + EWD               & 0.865    & 0.857    & 0.909      & 0.811 \\ \hline
MNB + TF-IDF            & 0.868    & 0.863    & 0.897      & 0.831 \\ \hline
LR + EWD                & 0.874    & 0.870    & 0.901      & 0.841 \\ \hline
LR + NER                & 0.906    & 0.905    & 0.900      & 0.912 \\ \hline
SVM + NER               & 0.918    & 0.918    & 0.912      & 0.924 \\ \hline
LR + TF-IDF             & 0.975    & 0.975    & 0.971      & 0.980 \\ \hline
SVM + TF-IDF            & 0.980    & 0.980    & 0.978      & 0.982 \\ \hline
BERT (Fine-Tuning)      & \textbf{0.987} & 0.984 & 0.986 & 0.982 \\ \hline
LSTM - w2v             & \textbf{0.987} & \textbf{0.987} & \textbf{0.987} & \textbf{0.986} \\ \hline
\end{tabular}
\end{center}
\end{table}

Our findings suggest that Word2Vec embeddings outperformed BERT embeddings in this particular task, likely due to the specialized language domain of scientific abstracts. Specifically, Word2Vec was better able to capture word context within sentences, while BERT focused on predicting masked tokens or generating subsequent sentences in the document. Additionally, while Word2Vec generated fixed-size embeddings for each word, BERT's embeddings were contextualized and varied based on the word's usage in the document.

\begin{figure}[b!]
    \centering
    \includegraphics[width=\linewidth]{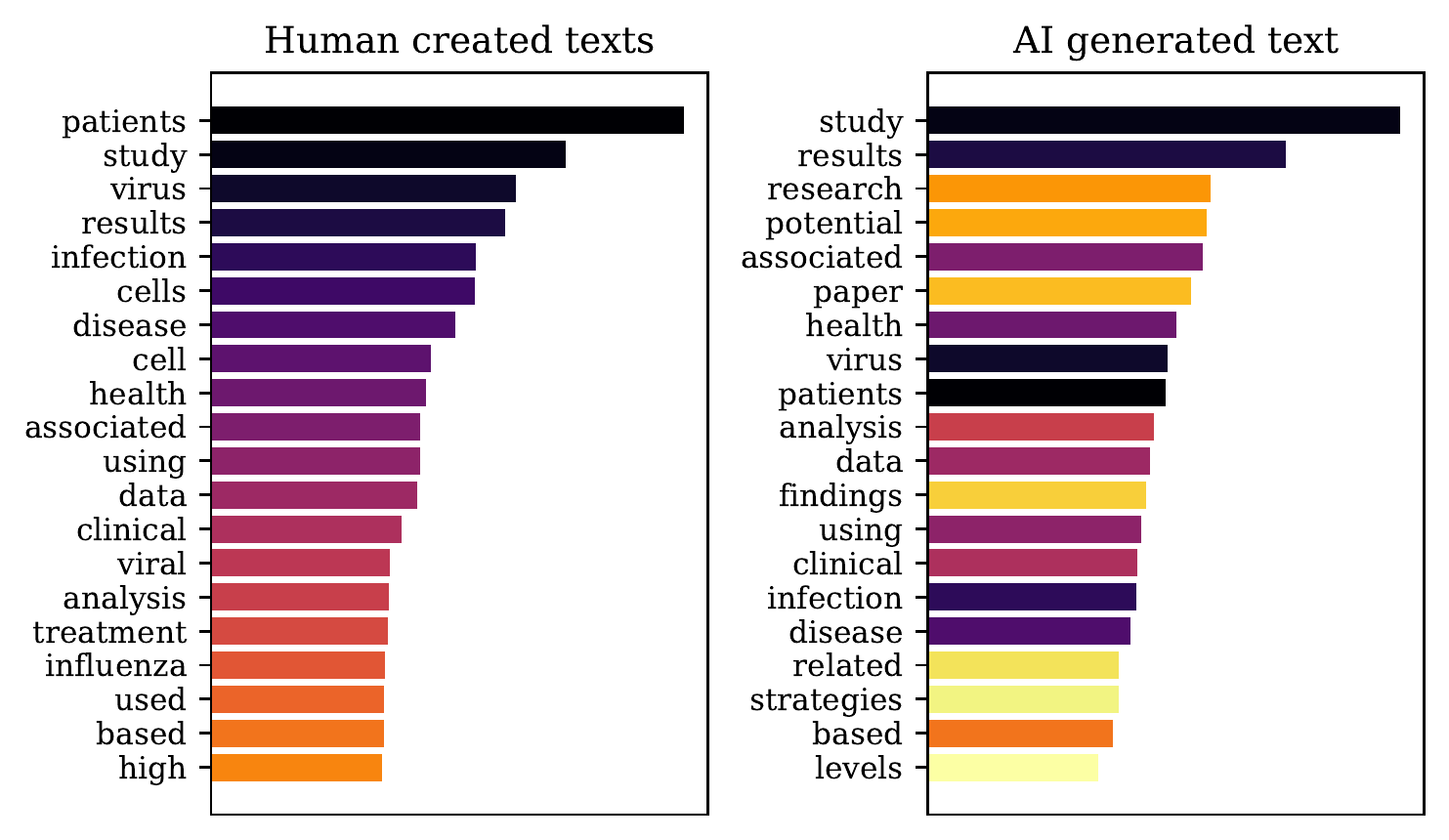}
    \caption{Most frequent word appearances in both the human-created texts (left) and the AI-generated texts (right).}
    \label{fig:word_freqs}
\end{figure}

\section{Discussion}\label{discussion}
The experimental results show that it can be detected the human created and AI-generated scientific abstracts highly accurate. The best model, LSTM with the Word2Vec representation) misclassified 103 samples from the total test samples, where 43 were human-created abstracts while the rest were AI-generated. It is obvious that the portion of misclassified texts is significantly low. Having set up the research question in Section~\ref{method} we continue with an explanatory analysis of the wrong classified abstracts, to understand the particularities of the AI-generated abstract that lead the model to classify them as human text. The crucial element of abstract construction is the title of the scientific papers, so we investigate the attributes of the titles.


\begin{figure}[t]
    \centering
    \includegraphics[width=\linewidth]{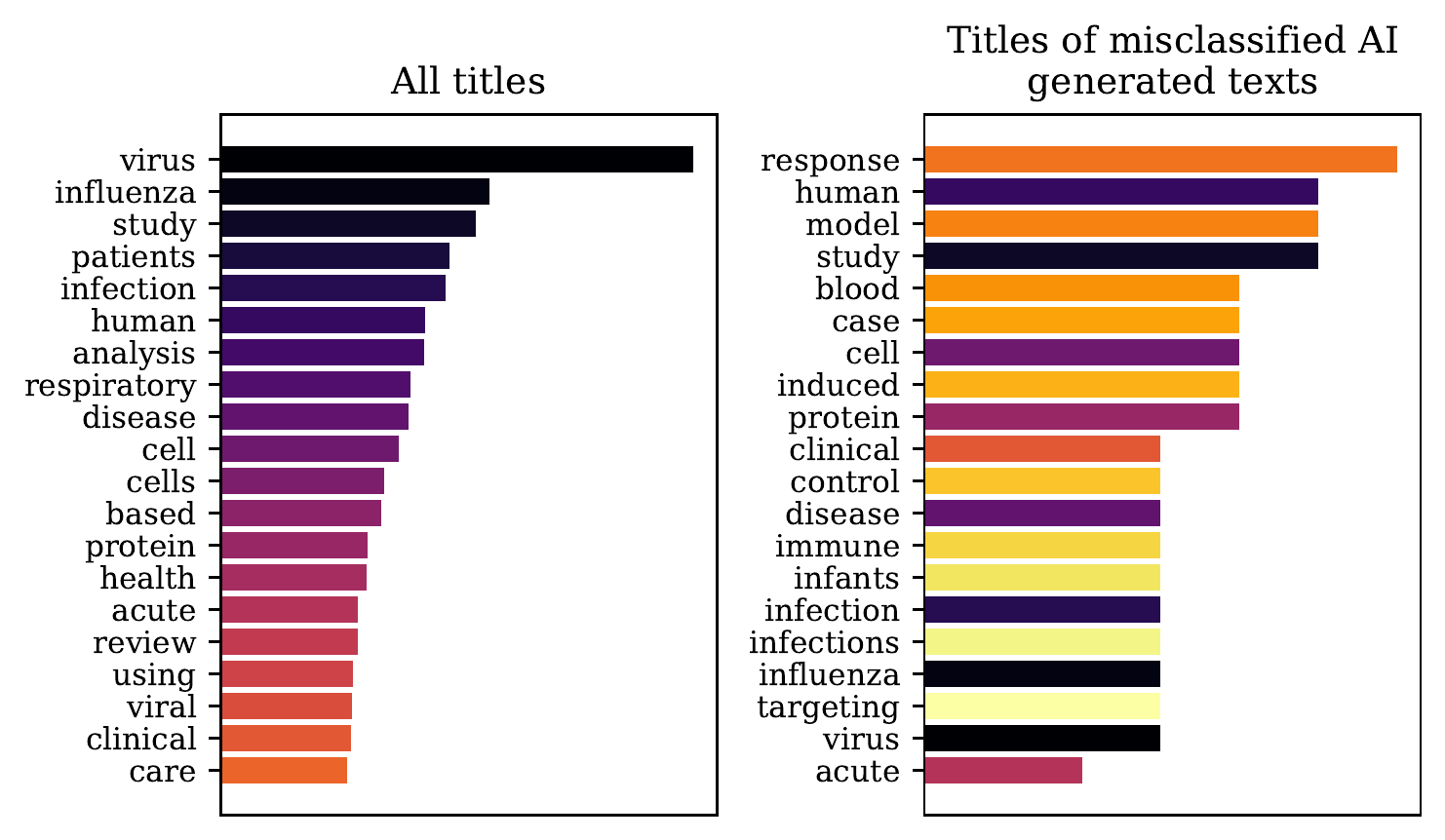}
    \caption{Most frequent word appearances in all the titles (left) and in the titles of the misclassified AI-generated titles (right).}
    \label{fig:ai_titles}
\end{figure}

\begin{figure}[b!]
    \centering
    \includegraphics[width=\linewidth]{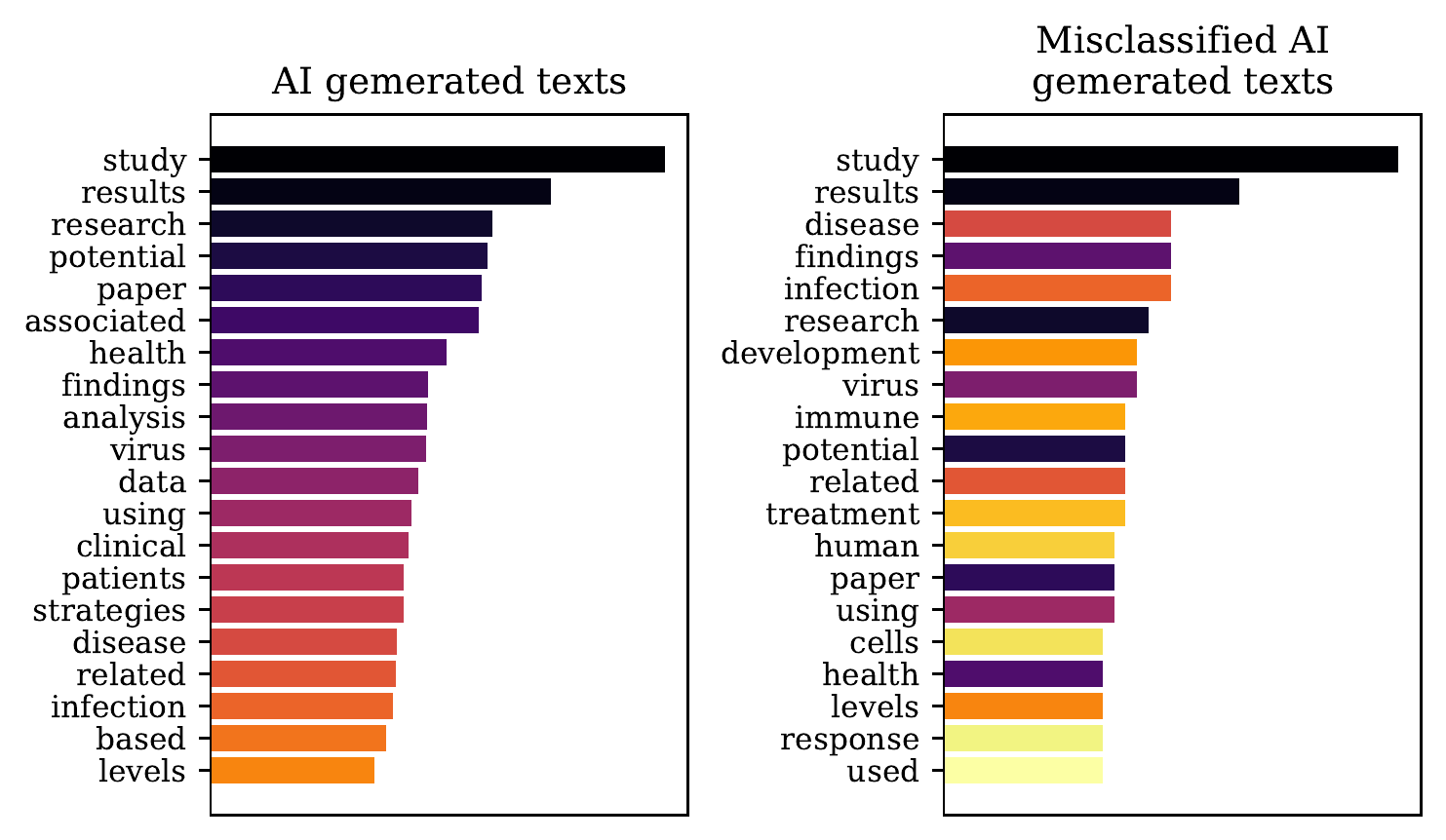}
    \caption{Most frequent word appearances in all the AI-generated abstracts (left) and in the misclassified AI-generated abstracts (right).}
    \label{fig:ai_abss}
\end{figure}


The first question we raised is, ``Does the size of the title we provided the GPT-3 model affect the classification error?'' The answer to this question is straightforward negative if we consider that the average size of the titles that constitute our dataset is 13.64 words, and the length of the misclassified generated text has an average title size of 13.68 words. At this point, there is no to pose the same question in the human-created texts.

The next question that needs answering is, ``What words dominate the misclassified AI-generated texts compared to the correctly classified ones?'' Before we answer this question, we need to understand the fundamental difference between human-created and AI-generated texts. To that end, Figure \ref{fig:word_freqs} presents the most frequently appeared words in both the human created and the AI-generated texts. It becomes immediately apparent from this figure that there is a difference between those two sets. In the human-created texts, we can see that they contain not-so-common words (such as \textit{influenza} or \textit{treatment}) with high frequency. On the contrary, AI-generated texts contain more widely used words. With this knowledge, we can now see Figure~\ref{fig:ai_abss} and Figure~\ref{fig:ai_titles}. The combination of those figures shows us that both the titles and the abstract of the misclassified AI-generated texts contain, with high frequency, not-so-common words, and in the case of the titles, they contain significantly more uncommon words than the rest of the titles of the dataset. Only from this fact can we conclude that the more information, in terms of content, contained in the titles given to the GPT-3 model, the better text generation it will conduct.

On the other hand, the human-created text that was misclassified had a poorer vocabulary than the main bulk of the human-created texts, so much so that the LSTM with Word2Vec methodology misclassified it as AI-generated text. The word analysis can be seen in Figure~\ref{fig:ai_abss}.

\begin{figure}[t]
    \centering
    \includegraphics[width=\linewidth]{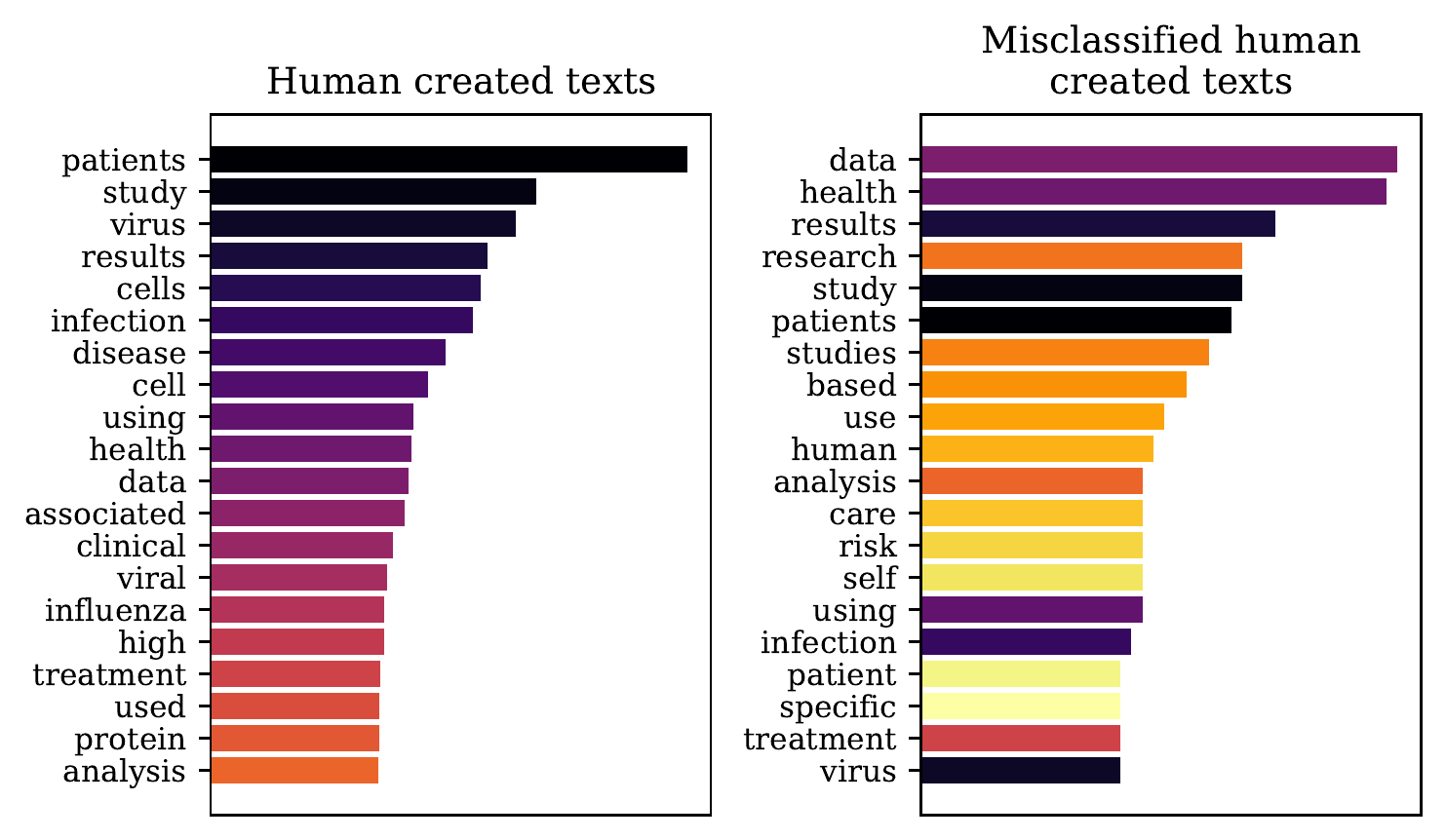}
    \caption{Most frequent word appearances in all the human-created abstracts (left) and in the misclassified human-created abstracts (right).}
    \label{fig:real_abss}
\end{figure}

\section{Conclusion}\label{conclusion}
In this work, we presented a method of distrusting between text generated by Artificial Intelligence and text created by humans. To that end, we presented several NLP classification methodologies, from simple ones, such as the LR with TF-IDF method, or more complicated ones, such as LSTM with BERT. From the results of the classification task, we can conclude that the problem at this point of the development of the GPT-3 model is seemingly a problem that can be 
tackled efficiently. 


Moving forward, we intend to produce a higher volume dataset using the updated state-of-the-art Large Language Model, with the aim of further evaluating and improving the efficacy of our proposed method. The larger dataset will enable us to investigate the generalizability and scalability of our approach across different domains, languages, and text genres. Moreover, it will allow us to explore the potential of our method for real-world applications. By advancing our understanding of the capabilities and limitations of AI-generated text, we can pave the way for more responsible and ethical use of this technology in the future.


\balance

\bibliographystyle{plain}
\bibliography{bibliography,bibl_panos}

\end{document}